\documentclass[letterpaper, 10 pt, conference]{ieeeconf}

\pdfoutput=1
\usepackage{hyperref}
\hypersetup{
	pdfinfo={
		Title={On the Hardware Feasibility of Nonlinear Trajectory Optimization for Legged Locomotion based on a Simplified Dynamics},
		Author={Angelo Bratta, Romeo Orsolino, Michele Focchi, Victor Barasuol, Giovanni Gerardo Muscolo and Claudio Semini},
		Subject={If you want to put something here, do so},
		Keywords={Add some keywords if you feel so inclined}
	}
}
\usepackage{pdfpages}

\IEEEoverridecommandlockouts                              

\usepackage{verbatim}
\usepackage{xcolor}
\usepackage{graphicx}
\usepackage{graphics} 
\usepackage{amsmath,amssymb}
\usepackage{bm}
\usepackage{bbold}
\usepackage{glossaries}
\usepackage{siunitx}
\usepackage{hyperref}

\newacronym{hyq}{HyQ}{Hydraulically actuated Quadruped}
\newacronym{cd}{CD}{Centroidal Dynamics}
\newacronym{srbd}{SRBD}{Single Rigid Body Dynamics}
\newacronym{dofs}{DoFs}{Degrees of Freedom}
\newacronym{com}{CoM}{Center of Mass}
\newacronym{mpc}{MPC}{Model Predictive Control}
\newacronym{haa}{HAA}{Hip Abduction-Adduction}
\newacronym{hfe}{HFE}{Hip Flexion-Extension}
\newacronym{kfe}{KFE}{Knee Flexion-Extension}

\newcommand{\eg}{\emph{e.g.,~}}
\newcommand{\etal}{\emph{et al.~}}
\newcommand{\ie}{\emph{i.e.,~}}

\newcommand{\Rnum}{\mathbb{R}} 

\newcommand{\vc}[1]{\mathbf{\bm{#1}}} 					
\newcommand{\mat}[1]{\ensuremath{\begin{bmatrix}#1\end{bmatrix}}}	

\title{\LARGE \bf On the Hardware Feasibility of Nonlinear Trajectory Optimization for Legged Locomotion based on a Simplified Dynamics}

\author{Angelo Bratta$^{1,2}$, Romeo Orsolino$^{1}$, Michele Focchi$^{1}$, Victor Barasuol$^{1}$,\\ Giovanni Gerardo Muscolo$^{3}$ and Claudio Semini$^{1}$
	\thanks{$^{1}$Dynamic Legged Systems (DLS) lab, Istituto Italiano di Tecnologia, Genova (Italy). email:{firstname.lastname}@iit.it}
	\thanks{$^{2}$ Department of Electronics and Telecommunications, Collegio di Ingegneria Informatica, del Cinema e Meccatronica, Politecnico di Torino, Turin (Italy).}%
	\thanks{$^{3}$Department of Mechanical and Aerospace Engineering, Politecnico di Torino, Turin (Italy). email: {giovanni.muscolo@polito.it}}%
}

\begin{document}
	
\null
\includepdf[pages=-]{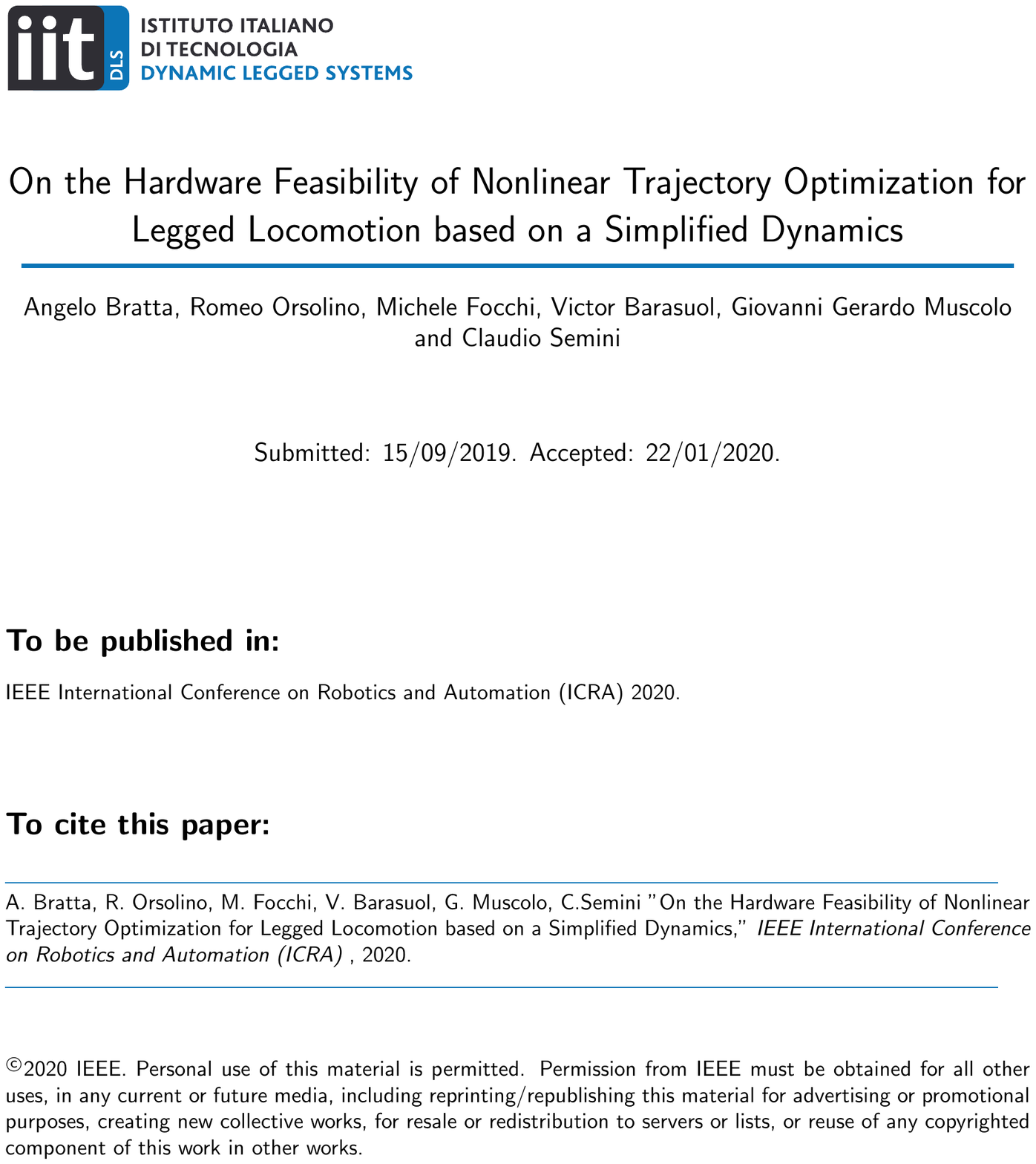}

\maketitle
\thispagestyle{empty}
\pagestyle{empty}
	
\begin{abstract}
Simplified models are useful to increase the computational efficiency of a motion planning algorithm, but their lack of accuracy have to be managed. 
We propose two feasibility constraints to be included in a Single Rigid Body Dynamics-based trajectory optimizer in order to obtain robust motions in challenging terrain. The first one finds an approximate relationship between joint-torque limits and admissible contact forces, without requiring the joint positions. The second one proposes a leg model to prevent leg collision with the environment. Such constraints have been included in a simplified nonlinear non-convex trajectory optimization problem. We demonstrate the feasibility of the resulting motion plans both in simulation and on the \gls{hyq} robot, considering experiments on an irregular terrain. 
\end{abstract}

\section{Introduction}\label{sec:intro}
	
Motion planning is a key element for the locomotion of legged robots. The more complex the terrain to be traversed the harder it gets to find a set of joint commands that allows the robot to reach a desired target.

The literature 
 is typically split between the usage of simplified dynamic models, such as the Linear Inverted Pendulum (LIP) \cite{Kajita2011} or the Spring Loaded Inverted Pendulum (SLIP) \cite{Geyer2018} and the usage of whole-body models. The first two are computationally efficient but only applicable to flat terrains; the latter guarantees 
 accurate description of the robot dynamics on arbitrary terrains but requires larger computational effort, 
 not suitable for real-time applications.

A third option is the \gls{cd} \cite{Orin2013}, \cite{Wensing2016} which exploits the full dynamics of a robot projected at its own \gls{com}. The \gls{cd} is not an approximate dynamic model, since it considers a complete description of the robot dynamics in terms of its inputs (feet and \gls{com} wrenches) and outputs (feet and \gls{com} trajectories). 

A simplification of the \gls{cd} is the \gls{srbd}, where the robot is seen as a single rigid body with massless legs. The leg inertia is constant and corresponds to a predefined joint configuration (what implies that the inertia effect of the leg is neglected). 

Due to its simplicity, \gls{srbd} is well suited to problems which require low computational cost while dealing with complex terrains and possible non-coplanar contacts. In addition, it is a suitable approximation for robots with negligible mass legs compared to the trunk's weight. 


Since \gls{srbd} only considers a description in terms of the \gls{com} and the robot's feet, it is possible that a motion planner based on this model finds a trajectory which violates feasibility constraints of the robot. Such constraints include the joint kinematic limits, the joint-torque limits and the possible collisions between robot's links and the environment. 


\subsection{Related Works}\label{sec:relatedWorks}

The trade-off between simplicity of the model and feasibility of a whole-body motion on rough terrain is still at the core of the research in robotic legged locomotion.
One approach consists in using the trajectory generated with \gls{cd} (which is typically lower dimensional than the full dynamic model of the robot) in combination with the full robot kinematics to obtain a feasible whole-body motion \cite{Dai2016d, Herzog2016}. A similar approach was used by Valenzuela \etal \cite{Valenzuela2016} in which the authors compute the sequence of footholds using a Mixed Integer Convex optimization problem and then solve a nonlinear problem to find a linear and angular centroidal momentum trajectory. Budhiraja \etal \cite{Budhiraja2019} generated a trajectory for the \gls{com}, in both simplified and full dynamic model, making sure that the former is coherent with the latter. All these approaches have been only tested in simulation with a bipedal robot, but never on the robotic hardware.

Farshidian \etal \cite{Farshidian2017} introduced a nonlinear \gls{mpc} which is able to compute the relevant motion quantities in a short amount of time. The limit of this approach is that it requires a big computational effort and it can thus only plan few phases of the motion.

Other authors leave on the motion controller the burden to verify the whole-body feasibility of the motion plan and focus, instead, on the nonlinearities which are present in the \gls{cd} and \gls{srbd} formulations (mainly angular momentum and contact phases duration). These are usually tackled by either defining a convex relaxation of the formulation \cite{Dai2016b, Ponton2018}, or by predefining the feet position for the entire trajectory \cite{DiCarlo2018a}. 


Winkler \etal \cite{Winkler2018} implemented a trajectory optimization, based on the \gls{srbd}, in which \gls{com} position, orientation, feet position, contact forces and timings are concurrently optimized. The efficiency of this planner has been demonstrated both in simulation and on the real hardware on flat terrains. On rough terrains, however, 
other feasibility constraints such as joint-torque limits and geometry of the robot's legs need to be considered to obtain a feasible whole-body motion suited for a real robotic hardware.

A state-of-art approach to avoid collision between legs and ground is looking for a collision-free swing phase, considering the height map of the terrain and the robot configuration \cite{Doshi2007}. An alternative is presented in \cite{Barasuol2019}, in which the controller is able to move
the point of application of the ground reaction forces from the foot to the shin in case of collision. This method can guarantee safe navigation on challenging terrains, however, it is a pure reactive module which does not increase the robustness of the planner.

The joint-torque limits constraint problem is usually only addressed at controller level \cite{Samy2017, Fahmi2019RAL}. In order to explicitly consider the limit at the planning stage, Ding \etal \cite{Ding2018} convexify the nonlinear joint-torque constraint such that it can be added to a Mixed Integer Quadratically Constrained Program (MIQCP). This formulation is suitable for convex optimization and it thus computationally efficient. The decision to employ a unique outer bounding ellipsoid as an approximation of the force ellipsoids, however, discards the important configuration-dependent information of the relationship between the leg's configuration and shape of the force ellipsoids.
In our previous works, instead, we have shown that the value of the maximum admissible contact forces depends on the leg's configuration and we have used the force polytopes \cite{Chiacchio1997} to map the joint-torques limits into a set of admissible centroidal wrenches \cite{Orsolino2018a} or \gls{com} positions \cite{Orsolino2018b}. This work employs the same idea of force polytope for the synthesis of motion planner algorithm based on \gls{cd}, or \gls{srbd} that retains a relationship between the robot configuration and the maximal contact forces.
	
\subsection{Contributions}
In this manuscript we address the limitations of the existing state-of-the-art nonlinear motion planners for legged robots based on \gls{srbd}. Our theoretical novelties allow to have a planner that devises contact forces consistent with actuation limits and feet trajectory that avoid shin collisions. Thanks to that, experimental results have been obtained for non-flat terrains. 
\subsubsection{Theoretical contributions}
\begin{itemize}
	\item a novel approximate projection of the joint-torque limits into the task space. This is limited to robots presenting point-contacts and, to the best of our knowledge, it represents a novel approach to describe the existing relationship between the leg's configuration and the corresponding maximal pure contact forces at the end-effector. This allows motion planners based on nonlinear optimization to adjust the robot configuration and forces depending on the manipulability and actuation capabilities of the platform
	\item a novel model of the leg's lower link to include into the trajectory optimization formulation the finite non-zero size of the robot's feet and shin's geometry. 
\end{itemize} 
\subsubsection{Experimental contribution} we present the hardware implementation of the trajectories obtained with a revisited version of the formulation presented by Winkler \etal \cite{Winkler2018}. The planner's trajectories have been tested on the \gls{hyq} \cite{Semini2011a} robot of Istituto Italiano di Tecnologia (IIT). This is the first time that trajectories based on \cite{Winkler2018} are deployed on a real hardware on non-flat terrains different from the flat ground, achievement which would not have been possible without the increased robustness guaranteed by the two feasibility constraints proposed in this paper. 

\section{Formulation of the problem}\label{sec:problemFormulation}
We attempt to overcome the already mentioned limitations by defining a motion planner which solves a nonlinear, non-convex, trajectory optimization problem based on the \gls{srbd} and which includes some nonlinear constraints which will bias the solver towards a solution that is more coherent with the whole-body model of the specific robot.

\subsection{Single Rigid Body Dynamics Model}
The assumption of configuration independent \gls{com} and inertia of the robot is a fair assumption especially for quadrupeds with legs of negligible mass compared to their trunk's mass, as in the case of \gls{hyq} \cite{Semini2011a}. For example, we have found that \gls{hyq}'s leg configuration changes the position of the \gls{com} of $2$cm in $x$ and $y$ direction at maximum, while the $z$ direction is less relevant for stability. The Newton-Euler equations lead to the following dynamic equation:
\begin{equation}
m\mathbf{\ddot{r}}=
\sum_{i=1}^{n_i} \mathbf{f}_i(t)-m\mathbf{g}
\label{DynamicModel1}
\end{equation}
\begin{equation}
\mathbf{I} \vc{\dot{\omega}}(t)+
\vc{\omega}(t)\times\mathbf{I} \vc{\omega}(t)=
\sum_{i=1}^{n_i} 
(\mathbf{f}_i(t)\times(\mathbf{r}(t)-\mathbf{p}_i(t)))
\label{DynamicModel2}
\end{equation}
where \textit{m} is the mass of the robot,
$\mathbf{r}$ $\in$ $\mathbb{R}^{3}$ is the CoM position
 $\mathbf{\ddot{r}}$ 
$\in$ 
$\mathbb{R}^{3}$ is the CoM linear acceleration, 
$\mathbf{f}_i$ 
$\in$
$\mathbb{R}^{3}$ are the pure contact forces, 
$n_i$ 
$\in$ $\mathbb{R}$ is the number of feet, 
$\mathbf{g}$
$\in$ $\mathbb{R}^{3}$ is the gravity vector, $\mathbf{I}$ $\in$ $\mathbb{R}^{3\times 3}$ is the inertia matrix, 
$\vc{\omega}$ $\in$ $\mathbb{R}^{3}$ is the angular velocity, $\vc{\dot{\omega}}$ $\in$ $\mathbb{R}^{3}$ is the angular acceleration, and $\mathbf{p}_i$ $\in$ $\mathbb{R}^{3}$ is the cartesian position of foot $i$.

\subsection{Kinematic Model}
In order to guarantee that joints lie inside their kinematic limits, the foot position with respect to the base is constrained to be inside a conservative approximation of the leg's workspace defined as a cube. This box has edges of size $2b$ and it is centered in the nominal foot position $\overline{\mathbf{p}}_i$ $\in$ $\mathbb{R}^{3}$ \footnote{The nominal foot position corresponds to a joint configuration which maximizes the distance of each joint from its kinematic limits.} :
\begin{equation}
\mathbf{p}_i(t) \in
\mathcal{R}_i
\Leftrightarrow
  |\mathbf{p}_i(t)- \overline{\mathbf{p}}_i|
  	<b
  	\label{KinematicModel}
\end{equation}
\subsection{Terrain and Contact Models}
This constraint enforces the fact that the feet have to be in contact with the terrain during the stance phase, and that they have to keep a minimum clearance $h_{min} \in \Rnum$ from the terrain during swing:

\begin{equation}
p_z \in \mathcal{T} 
\Rightarrow
\begin{cases}
p_z = h_{ter}(p_x, p_y) & \text{during stance} \\
p_z > h_{min} + h_{ter}(p_x, p_y) & \text{during swing}
\end{cases}
\end{equation}

where $p_z$ is the $z$ coordinate of the foothold and $h_{ter}$ is the height of the ground obtained through a predefined internal terrain model.

The contact model of \cite{Winkler2018} checks whether the contact force $\vc{f}_i$ of every foot lies within the linearized friction cone $\mathcal{F}_i$. From a mathematical point of view it has to verify that:
\begin{equation}
\underline{\vc{a}}_i<\vc{C}_i\vc{f}_i<\bar{\vc{a}}_i
\end{equation}
%
\begin{equation}
\begin{aligned}
\vc{C}_i=
\begin{bmatrix}
(-\mu_i\vc{s}_i+\vc{t}_{1i})^T\\
(-\mu_i\vc{s}_i+\vc{t}_{2i})^T\\
(\mu_i\vc{s}_i+\vc{t}_{2i})^T\\
(\mu_i\vc{s}_i+\vc{t}_{1i})^T\\
\vc{s}_i^T\\
\end{bmatrix}
\underline{\vc{a}}_i =
\begin{bmatrix}
-\infty\\
-\infty\\
0\\
0\\
0\\
\end{bmatrix}
\bar{\vc{a}}_i =
\begin{bmatrix}
0\\
0\\
\infty\\
\infty\\
f_{max}
\end{bmatrix}
\end{aligned}
\end{equation}

where $\mathbf{s} \in \Rnum^3$ is the normal direction to the terrain, $\mu$ is the friction coefficient,  $\mathbf{f}_i \in \Rnum^3$ is the contact force, $\mathbf{t}_{1i}$, $\mathbf{t}_{2i}$ are the two tangential directions to the terrain, they form with $\mathbf{s}$ a right handed system of coordinates and $i = 1, \dots n_c$ where $n_c$ is the number of feet in contact with the ground.

In this paper, we use a formulation which corresponds to the one employed by Winkler \etal \cite{Winkler2018}, adding the \textit{force polytope} $\mathcal{A}_i$ and the \textit{leg's geometry} $\mathcal{P}_i$ constraints that are the contribution of this work and are further explained in the following Section.
The overall trajectory optimization problem can be described as:
\begin{equation}\label{eq:formulation}
\begin{aligned}
\text{find: } 
\quad &  \vc{x}(t) = 
[\vc{r}^T(t),  \vc{\theta}^T(t), \vc{p}_i^T(t), \vc{f}_i^T(t)] \\
& \text{for} \quad i = 0, \dots n_c\text{,} 
 \ \quad t=0,dT, 2dT, \dots T_f\text{,}\\
& \text{with} \quad dT=0.1 s \quad \text{and} \quad T_f= \text{final time}\\
\text{such that:}\\
(\text{dynamic model}) & \quad [\ddot{\vc{r}}, \dot{\vc{\omega}}]^T = f(\vc{r}, \vc{\theta}, \vc{p}_i,\vc{f}_i)\\
(\text{friction cone}) & \quad \vc{f}_i \in \mathcal{F}_i\\
(\text{foot position}) &	\quad     \vc{p}_i \in \mathcal{R}_i 
\cup \mathcal{T}_i\\
{(\text{force polytope})} &	\quad     {\vc{f}_i \in \mathcal{A}_i}\\
{(\text{leg's geometry})} &	\quad     {\vc{p}_i \in \mathcal{P}_i}\\
\end{aligned}
\end{equation}
where $\vc{\theta} \in SO(3)$ is the orientation of the robot.
\section{Geometry and Actuation Consistency}\label{sec:feasibilityConstraints}
In this Section, we propose a method to incorporate the joint-torque limits constraint and shin collision avoidance into simplified models such as \gls{cd} and \gls{srbd}.

\subsection{Joint-Torques Limits}\label{sec:torqueLimits}
\begin{figure}
		\centering

		\includegraphics[width=6cm]{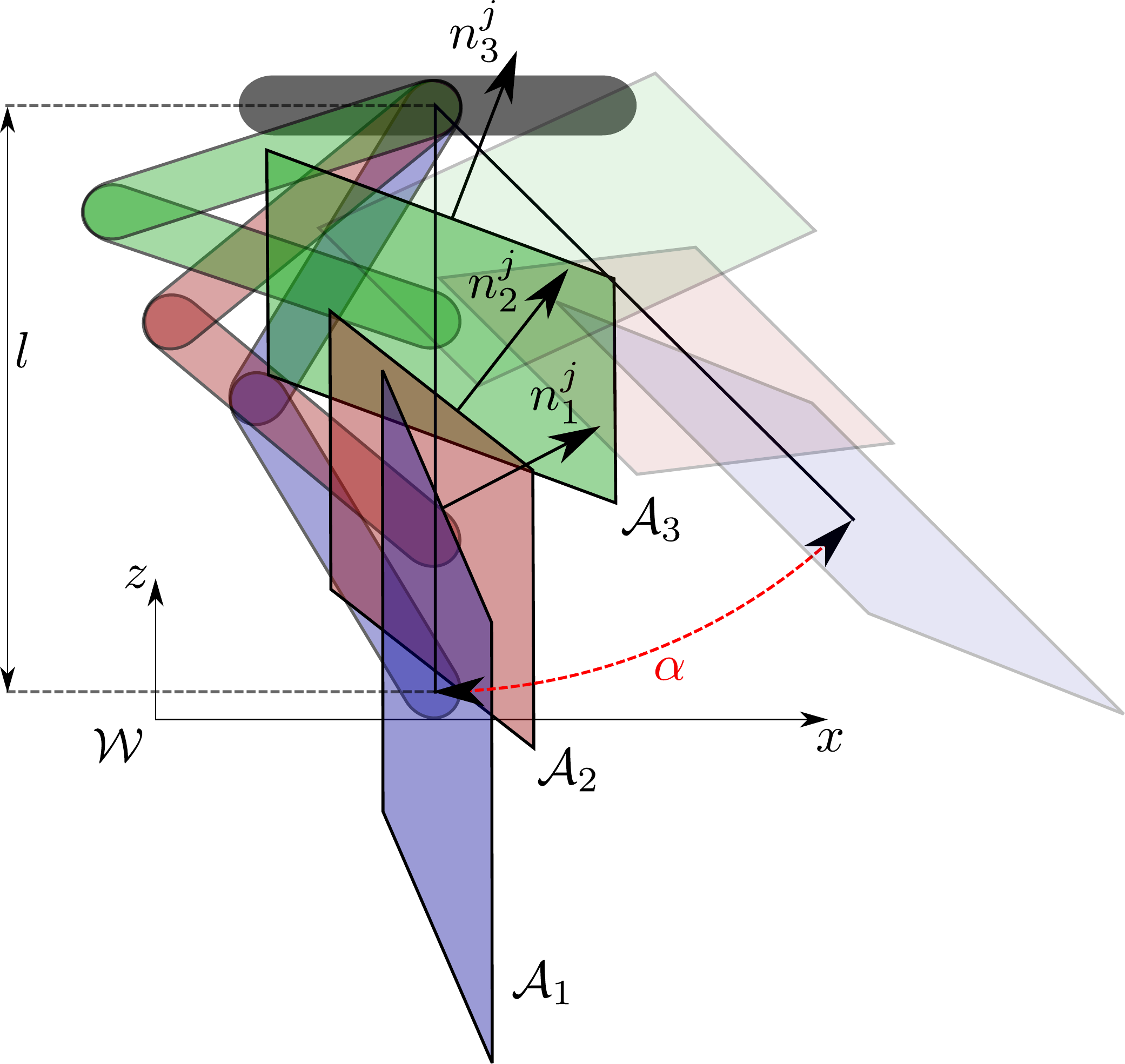}

		\caption{Representation of a planar leg with 2 \gls{dofs} in three different configurations. For each configuration the corresponding force polytope is shown. The angle $\alpha$ represents the tilt angle of leg (\ie angle between the vertical and the line connecting the foot and hip joint). $n_k^j$ is the normal to the j-th plane of the k-th polytope. $l$ is the distance between foot and hip joint. Shadow zones represent the polytopes when $\alpha$ is different from zero.}
		\label{fig:polyMorphing}
	\end{figure}

Traditional contact models neglect the existing relationship between the maximum contact force $\vc{f}_i^{lim}$ and the limb configuration $\vc{q}_i \in \Rnum^{n_d}$ (where $n_d$ is the number of \gls{dofs} of the leg). On the robotics hardware, admissible joint-torques $\vc{\tau}_i(\vc{q})$ are mapped into admissible contact forces $\vc{f}_i$ at the $i^{th}$ end-effector \cite{Samy2017, Chiacchio1997, Orsolino2018a} (from now on, unless specified, all the quantities refer to one single leg and we can thus drop the pedex $i$). This mapping is represented by the leg's Jacobian $\vc{J}(\vc{q})$ matrix which is typically nonlinear and configuration dependent:
\begin{equation} \label{eq:tau}
\vc{\tau}^{lim} = \vc{J}(\vc{q})^{T} \vc{f}^{lim}
\end{equation}
where $\vc{f}^{lim}$ is one of the $2^{n_d}$ vertices of the leg's force polytope $\mathcal{A}$ \cite{Siciliano2007}. Equation (\ref{eq:tau}) represents a static relationship, which can be also considered as a good approximation for dynamic movements of robots with low inertia legs. For a maximum joint-torque value $\vc{\tau}^{lim}$, the polytope $\mathcal{A}$ can be employed to relate the feet's position to the maximum contact forces that the robot can perform on the ground. This is an important feature that needs to be considered whenever the robot needs to take on complex configurations in order to negotiate rough terrains. 
In Fig. \ref{fig:polyMorphing} we can see a planar example where three force polytopes $\mathcal{A}_k$ (with $k = 1, \dots, 3$) are computed for the three default configurations of the same planar leg, corresponding to the situation of minimum retraction, of maximum extension and to the leg's nominal configuration. The selection of the number of default polytopes, three in this case, affects the precision of the approach: the higher the number of default sampled polytopes, the higher the accuracy of our approach since the resulting polytopes will be closer to the precomputed one. 
The considered leg has 2 \gls{dofs} and, therefore, all the force/wrench polytopes are made of $2n_d$ halfspaces. Each halfspace can be represented by its normal unit vector $\vc{n}_k^j \in \Rnum^m$ (with $j = 1 \dots 2n_d$) and the offset term $d_k^j \in \Rnum$. The force polytope $\mathcal{A}_k$ of (\ref{eq:formulation})  can be expressed as:	
\begin{equation}
	\mathcal{A}_k = \Big\{ \vc{f} \in \Rnum^m \quad | \quad \vc{A}_k(\vc{q}) \vc{f} \leq \vc{d}_k(\vc{q}) \Big\}
	\label{eq:staticForcePolyInequality}
\end{equation}
where $m$ is the dimension of the contact force (\ie$m=3$ or $m=2$ as in the planar case of Fig. \ref{fig:polyMorphing}) and:
\begin{equation}
\begin{aligned}
	&\vc{A}_k(\vc{q}) = \mat{\vc{n}_k^1(\vc{q}), \hdots, \vc{n}_k^{2n_d}(\vc{q})}^T \\
	&\vc{d}_k(\vc{q}) = \mat{d_k^1(\vc{q}), \hdots, d_k^{2n_d}(\vc{q})}^T
\end{aligned}
\end{equation}
The rows of the $\vc{A}_k(\vc{q})$ matrix are related to the rows of the leg's transposed Jacobian $\vc{J}(\vc{q})^T$. As we are interested in the way the matrix $\vc{A}_k(\vc{q})$ changes with respect to a variation of the foot $\vc{p}$ in the cartesian space, we should then analyze the quantity $d \vc{J}(\vc{q}) / d \vc{p}$. This is, however, robot-specific and goes against the assumption of \gls{srbd}. For this reason, we then propose a robot-agnostic approximate approach which expresses the relationship between foot position and maximal contact forces without the need of explicitly knowing the morphology of the considered robot.

In order to express the position of the foot we use the polar coordinates with respect to the hip joint in the sagittal plane. In particular,  the variable $l$ represents the distance between the foot and the hip joint and $\alpha$ corresponds to the angle between the vertical and the line that connects the foot to the hip joint (see Fig. \ref{fig:polyMorphing}) :
\begin{equation} \label{eq:l}
l = \sqrt{({}_Bp_x - {}_Bh_x)^2 + 
	({}_Bp_z - {}_Bh_z)^2 }
\end{equation}
\begin{equation} \label{eq:alpha}
	\alpha = \arctan \Big(\frac{{}_Bp_x - {}_Bh_x}{{}_Bp_z - {}_Bh_z} \Big)
\end{equation}
where $({}_Bh_x, {}_Bh_z)$ is the fixed position of the hip joint and $({}_Bp_x, {}_Bp_z)$ is the foot position with respect to the base frame.
$l_1, l_2$ and $l_3$ correspond to the distance between hip and foot, respectively, at maximum retraction, nominal position and at maximum extension.
In the following notation, the indexes $(\cdot)_a$ and $(\cdot)_b$ refer, for a given foot distance $l$, to the values corresponding to the two neighboring force polytopes $\mathcal{A}_a$ and $\mathcal{A}_b$ out of the three predefined polytopes $\mathcal{A}_1, \mathcal{A}_2, \mathcal{A}_3$:
\begin{figure}[t]
	\centering
	\includegraphics[width=6cm]{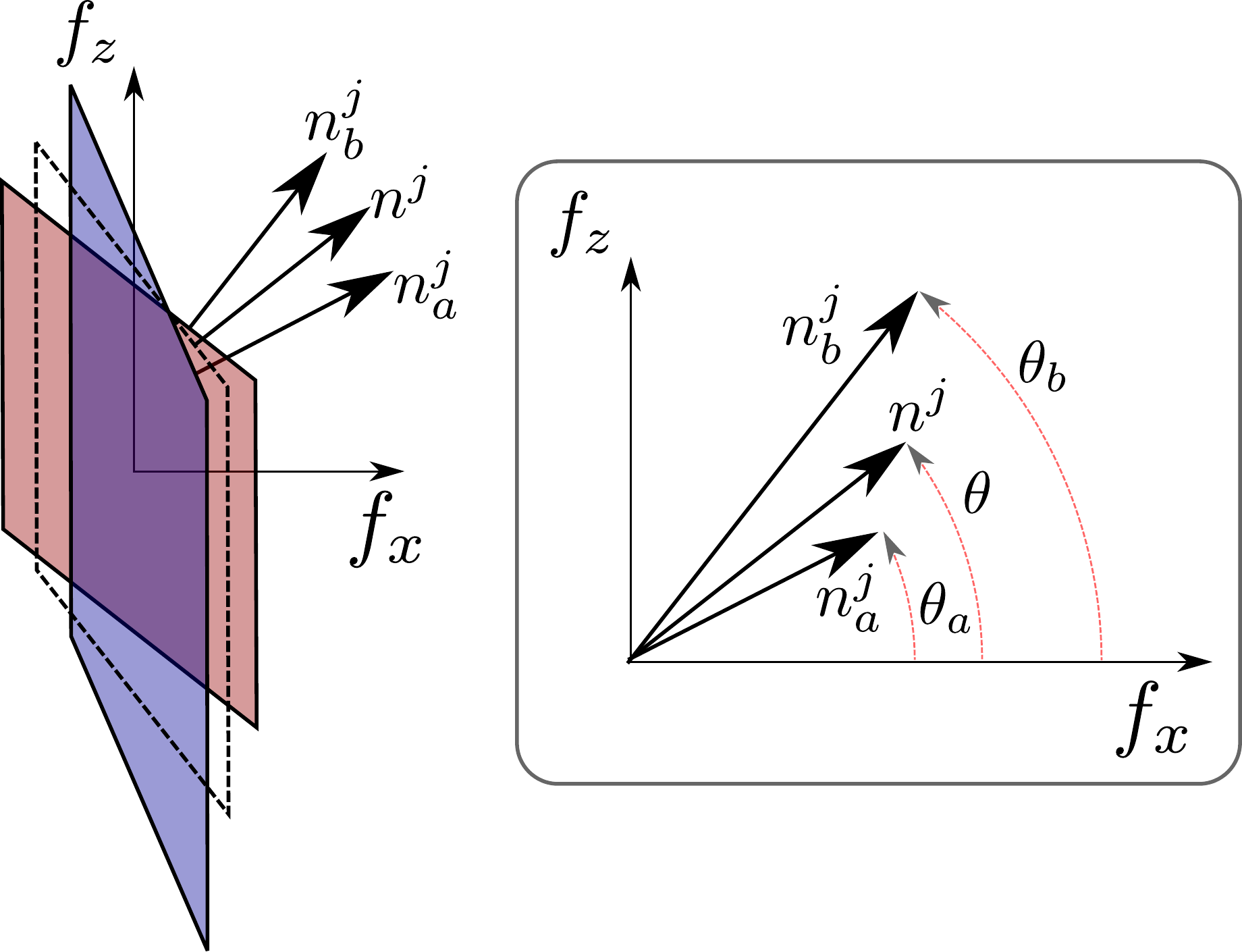}
	\caption{Geodesic average of the unit vectors $\vc{n}_a^j$ and $\vc{n}_b^j$ normal to the halfplanes of the force polytopes $\mathcal{A}_a$ and $\mathcal{A}_b$. The generic normal vector $\vc{n}^j$ is obtained by linear interpolation of the angles $\theta_a$ and $\theta_b$.}
	\label{fig:geodesicAverage}
\end{figure}
\begin{equation}
\Bigg\{ 
\begin{aligned}
(\cdot)_a = (\cdot)_1^j, \quad (\cdot)_b = (\cdot)_2^j \quad \text{if} \quad l_1 \leq l \leq l_2 \\
(\cdot)_a = (\cdot)_2^j, \quad (\cdot)_b = (\cdot)_3^j \quad \text{if} \quad l_2 < l \leq l_3
\end{aligned}
\label{eq:interpolation}
\end{equation}
We assume that the transformation between the corresponding halfplanes of two polytopes $\mathcal{A}_a$ and $\mathcal{A}_b$ consists of an homogeneous transformation (polytope morphing). 
The generic normal unit vector $\vc{n}^j(\vc{p})$ (describing the orientation of the force polytope's facet $j$) is found as the geodesic average \cite{Gramkow2001} of the two neighboring values $l_{a}$ and $l_{b}$:
\begin{equation}
\vc{n}^j(\vc{p}) = \vc{R}(\alpha) \mat{\cos(\theta) \sin(\theta)}^T
\label{eq:normalVector}
\end{equation}
where:
\begin{equation} 
\theta = \frac{l - l_{a}}{l_{b} - l_{a}}(\theta_b - \theta_a) + \theta_a
\label{eq:theta}
\end{equation}
$\theta_a$ and $\theta_b$ are the angles corresponding to the two \textit{predefined} normal vectors $\vc{n}_a = [\cos(\theta_a), \sin(\theta_a)]^T$ and $\vc{n}_b = [\cos(\theta_b), \sin(\theta_b)]^T$ closest to the value of $l$ (see Fig. \ref{fig:geodesicAverage}). The rotation matrix $\vc{R}(\alpha) \in \Rnum^{2\times2}$ maps the obtained force polytope by the angle $\alpha$ in such a way to align it to the leg. 
The generic offset term ${d}^j$ can be found as a linear interpolation between the values ${d}_a$ and ${d}_b$: 
\begin{equation}
d^j(\vc{p}) = \frac{l - l_{a}}{l_{b} - l_{a}}(d_b - d_a) + d_a
\label{eq:knownTerm}
\end{equation}
As a summary, following (\ref{eq:l}) - (\ref{eq:knownTerm}) we obtain an expression of the polytope which does not depend on the joint configuration $\vc{q}$, but on the foothold $\vc{p}$. The optimization problem \ref{eq:formulation} is fed a force polytope for each contact foot $i$:
\begin{equation}
\mathcal{A} = \Big\{ \vc{f} \in \Rnum^m \quad | \quad \vc{A}(\vc{p}) \vc{f} \leq \vc{d}(\vc{p}) \Big\}
\end{equation}
\begin{equation}
\begin{aligned}
&\vc{A}(\vc{p}) = \mat{\vc{n}^1(\vc{p}), \hdots, \vc{n}^{2n}(\vc{p})}^T \\
&\vc{d}(\vc{p}) = \mat{d^1(\vc{p}), \hdots, d^{2n}(\vc{p})}^T 
\end{aligned}
\end{equation}

Despite the nonlinearity given by the trigonometric terms in (\ref{eq:normalVector}), the polytope morphing of (\ref{eq:normalVector}), (\ref{eq:theta}) and (\ref{eq:knownTerm}) represents a significant simplification because it does not require the knowledge of the leg's Jacobian matrix $\vc{J}(\vc{q})$ at the considered leg configuration. This morphing is well suited for trajectory optimization formulations such as the one described in (\ref{eq:formulation}) where the optimization variables only includes operational space quantities.

\subsection{Environment Collision Avoidance}\label{sec:shinCollision}
\begin{figure}
	\centering
	\includegraphics[width=6cm]{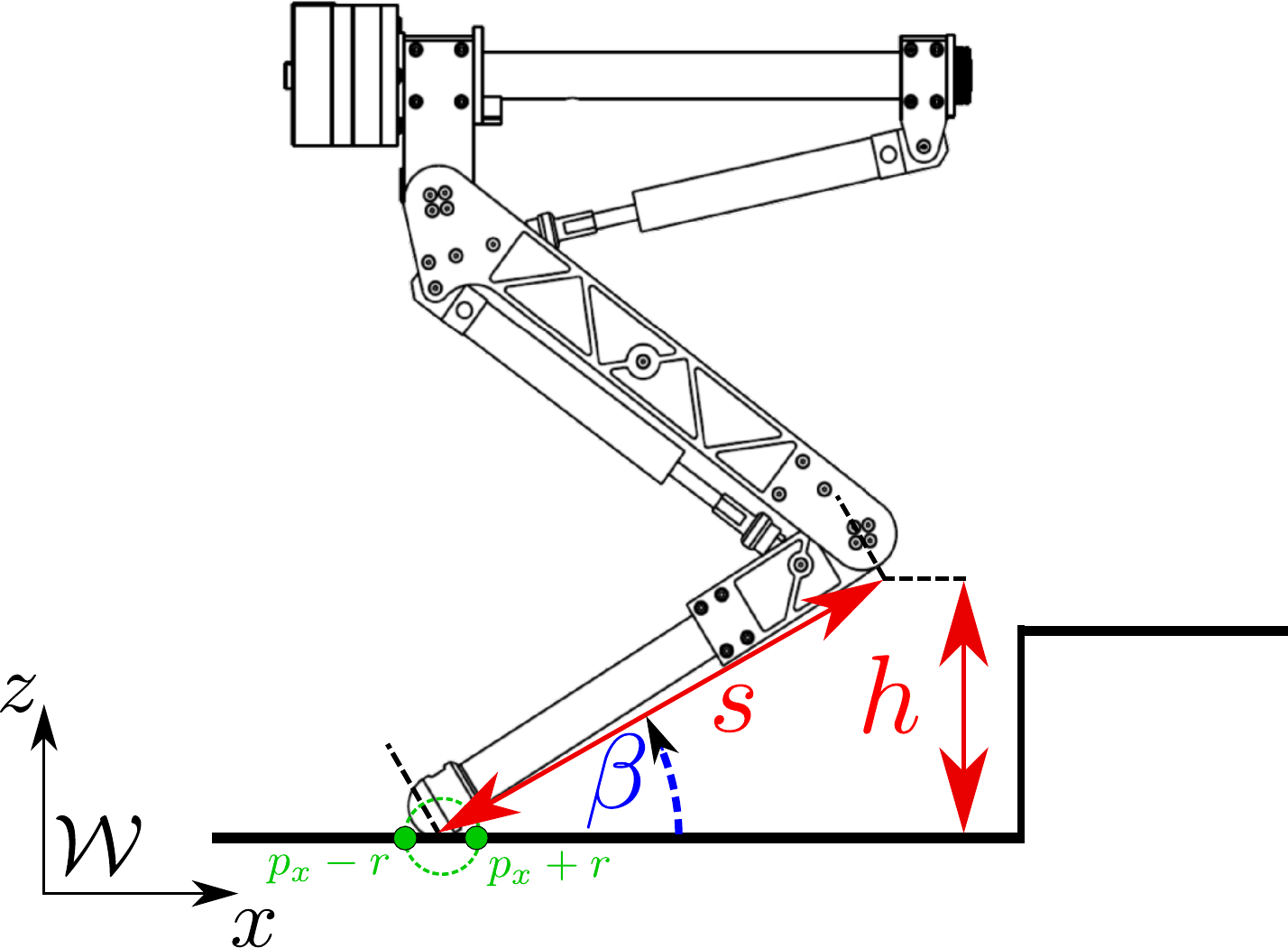}
	\caption{Representation of one single leg of the \gls{hyq} robot. $h$ is the knee height, $s$ is the length of the lower link, $\beta$ is the angle selected to keep the shin from hitting possible obstacles and $r$ is the radius of the foot used to avoid edges and unsafe foothold.}
	\label{fig:shinCollision}
\end{figure}
As introduced in Section \ref{sec:intro}, the \gls{srbd} considers the assumption of point feet, while in the real robots feet have always a non-negligible size. In this Section, we include the radius of robot's spherical feet in the formulation in such a way to discard solutions that may lead to undesired collisions during swing phase or to unsafe footholds in presence of obstacles and rough terrains.
Unsafe footholds, for example, happen when the spherical foot does not step entirely on the terrain (\eg edge of a step). To overcome this, we force the planner to find a foothold $\vc{p}$ in which the terrain height is constant at a radius $r$ before and behind the considered foothold along the robot's direction of motion (see Fig. \ref{fig:shinCollision}). The value of $r$ coincides with the physical dimension of the ball foot of the robot, \eg$2$cm in the case of \gls{hyq}.

Moreover, the leg's volume plays an important role in the execution of successful motion plans, as it may lead to self-collisions or to collisions with the environment if not properly managed. Shin collisions, for example, may either occur during a leg's stance phase (as a consequence of a poor choice of the foothold) or during a leg's swing phase. In order to avoid such collisions, we provide a simplified kinematic model of the leg to the motion planner. We assume the lower link to be a straight line of length $s$ (distance between the foot's contact point and the knee) with a fixed angle $\beta$ with respect to the ground. The knee's projection on the ground is equal to $s \cos(\beta)$ and the height of the knee corresponds to $h=s \sin(\beta)$ and it can then be mapped along the direction of motion using the knowledge of the yaw angle $\psi$ of the robot.  
Knee collision can thus be avoided if the height of the knee is higher than the terrain on that point. 
Fig. \ref{fig:shinCollision} represents the model chosen for the $i^{th}$ leg in the x-z plane:
\begin{equation} \label{eq:shin_collision}
\vc{p} \in \mathcal{P} \Rightarrow \quad
p_z + s \cdot \sin(\beta) >
h_{ter}(\vc{p}_{knee})
\end{equation}
\begin{equation}
\begin{aligned}
p_{knee,x}=
p_x &+ s \cos(\beta) \cos(\psi) \\
p_{knee,y}=
p_y &+ s \cos(\beta) \sin(\psi)
\label{knee_position_y}
\end{aligned}
\end{equation}
$s$ is a constant robot parameter while the value of $\beta$ angle is selected by looking at the maximum inclination of the leg 
during a walk.
Besides checking for possible knee collisions, we also avoid shin collisions by imposing (\ref{eq:shin_collision}) on a number of points between the foot and the knee. The number of these points is selected through heuristics and it depends on the length of the leg and on the roughness of the terrain. 

\section{Simulation and Experimental Results}\label{sec:results}

In this Section we present the validation results that have lead to the successful execution of the optimal trajectories generated by our motion planner on the \gls{hyq} robot, both in simulation and on the real hardware. For all the simulations and experiments we used an Intel® Core™ i5-4460 CPU @ 3.20GHz $\times$ 4 and the nonlinear optimization problem was solved using an Interior Point method \cite{Yamashita1998} solver, implemented in the IPOPT library \cite{Wächter2006}.
 
\subsection{Joint-Torque Limits Approximation}

\begin{figure}
	\centering
	\includegraphics[width=\linewidth]{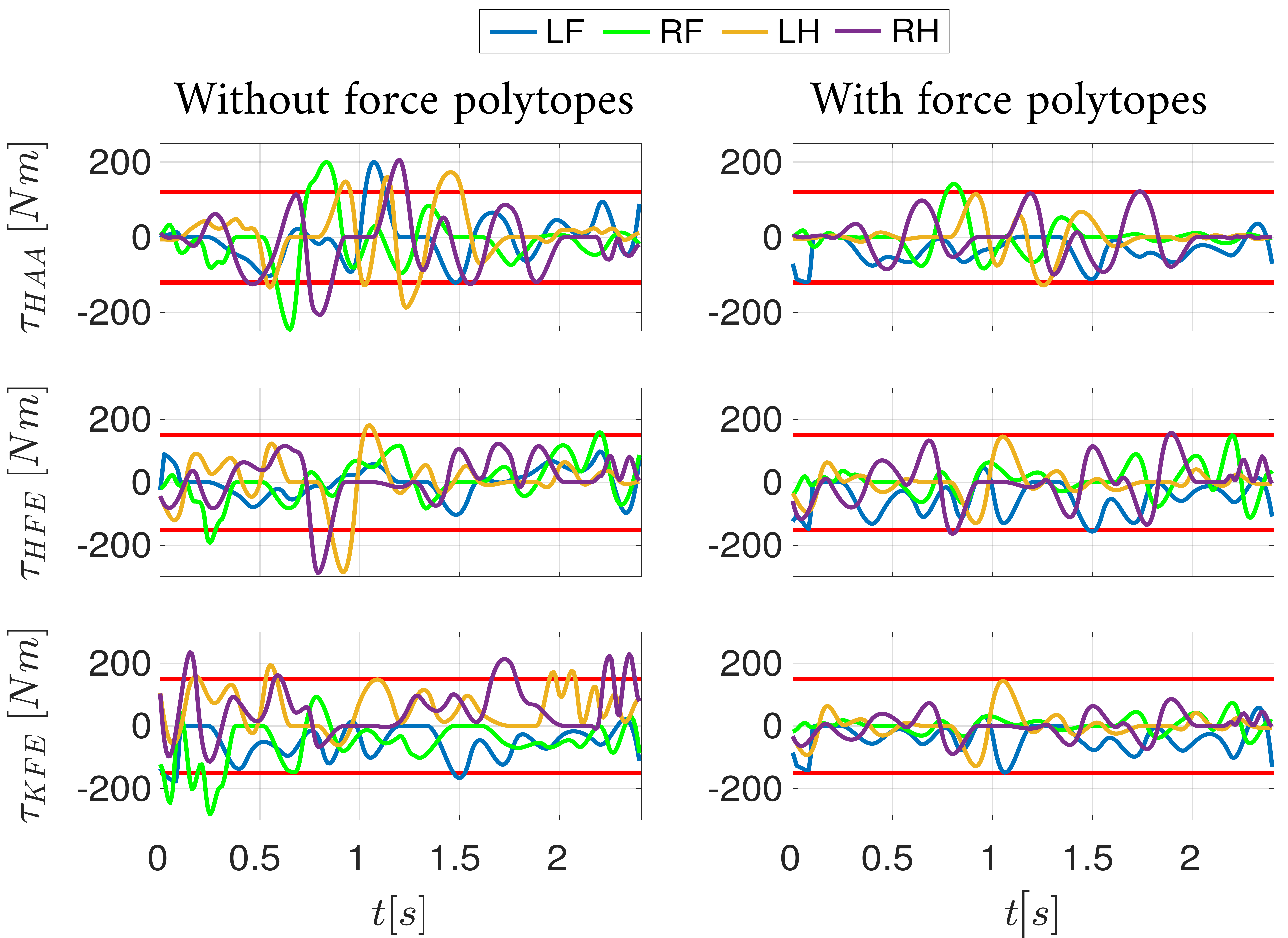}
	\caption{This plot shows the joint-torques of the \gls{hyq} robot walking in simulation for $2.4$s on a flat terrain. We can see, in the case where no force polytopes are considered (left plots), that the torques $\tau_{HAA}, \tau_{HFE}$ and $\tau_{KFE}$ violate their limits multiple times during the walk. On the right plots, instead, we can see that the force polytope constraint is able to bias the planner towards a solution that respects all the limits. Minor violation at $t = 0.8$s ($\tau_{HAA,RF}$ and $\tau_{HFE,RH}$).}
	\label{fig:torqueLimitsMarginHyq}
\end{figure}

Fig. \ref{fig:torqueLimitsMarginHyq} shows the \gls{haa}, \gls{hfe} and \gls{kfe} joint-torques and the corresponding saturation limits of the \gls{hyq} robot during a $1$m walk on a flat terrain (duration $2.4$s - three crawl gait cycles).

The plots show the torques obtained using the motion planner with (right) and without (left) the force polytope constraints. Our motion planner does not explicitly optimize over joint-torques $\vc{\tau}$ and so they are obtained exploiting the dynamic equation of motion of each single leg \footnote{ \gls{hyq}'s control structure is equipped with a whole-body controller to compute joint-torques, see \cite{Fahmi2019RAL} for further information.} :
\begin{equation}
\vc{\tau} = \vc{M} \vc{\ddot{q}} +  \vc{c}(\vc{q}, \dot{\vc{q}}) + \vc{g}(\vc{q}) -  \vc{J}(\vc{q})^T \vc{f}
\end{equation}
$\vc{f}$ is the contact force as optimized by the motion planner and $\vc{q}, \dot{\vc{q}}, \ddot{\vc{q}}$ can be obtained by inverse kinematics of the foot trajectory in the base frame ${}_B\vc{p}$ (assuming a fixed offset between base and robot's \gls{com}). $\vc{M}, \vc{c}$ and $\vc{g}$ represent
the leg's inertial matrix, the Coriolis and the gravity term.

We can see in the left plots that the desired torques violate the saturation limits of the actuators of the \gls{hyq} robot (\SI[inter-unit-product =\ensuremath]{120}{\newton\meter} for the HAAs and \SI[inter-unit-product =\ensuremath]{150}{\newton\meter} for the HFEs and KFEs). This is justifiable considering that the baseline motion planner has no information about saturation values and only linearized friction cones constraints act on the contact forces.
\begin{figure}
	\centering
	\includegraphics[width=\linewidth]{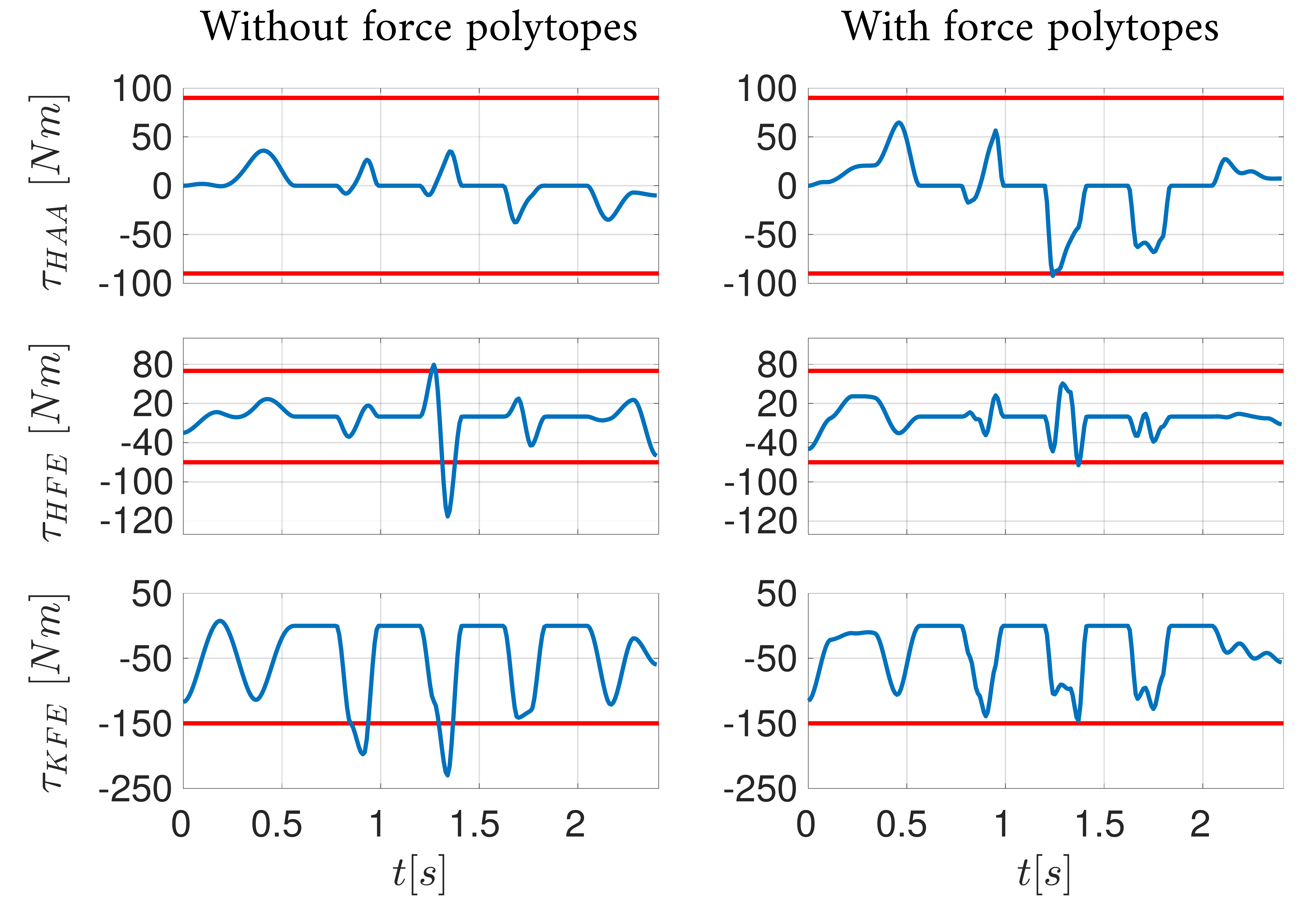}
	\caption{This plot shows the joint-torques of the monoped robot over a $2.4$s hopping on a flat terrain. We can see, in the case where no force polytopes are considered (left plots), that the HFE torque $\tau_{HFE}$ and $\tau_{KFE}$ limit are violated during the stance phases. 
	On the right plots, instead, we can see that the force polytope constraint is able to bias the planner towards a solution that globally respects all the limits. Minor violations at $t=1.3$s ($\tau_{HAA}$) and at $t=1.4$s ($\tau_{HFE}$).}
	\label{fig:torqueLimitsMarginMonoped}
	\vspace{-0.45cm}
\end{figure}
The plots on the right, instead, satisfy the torque limits of the robot due to the force polytope constraint included in the motion planner. This is possible thanks to more extended leg-configuration that \gls{hyq} takes on during the walk and the stance phases. As a matter of fact, such extended configuration corresponds to a force polytope with a larger maximum normal force (see Fig. \ref{fig:polyMorphing}) \cite{Orsolino2018b}. Minor violations happen, since the geodesic average is a good approximation of the polytope and not its real representation. Figure \ref{fig:torqueLimitsMarginMonoped} shows similar results for the monoped robot (corresponding to a single leg of \gls{hyq}, see Fig. \ref{fig:shinCollision}).

\subsection{Shin Collision Avoidance}
\begin{figure}[t]
	\centering
	\includegraphics[width=\linewidth]{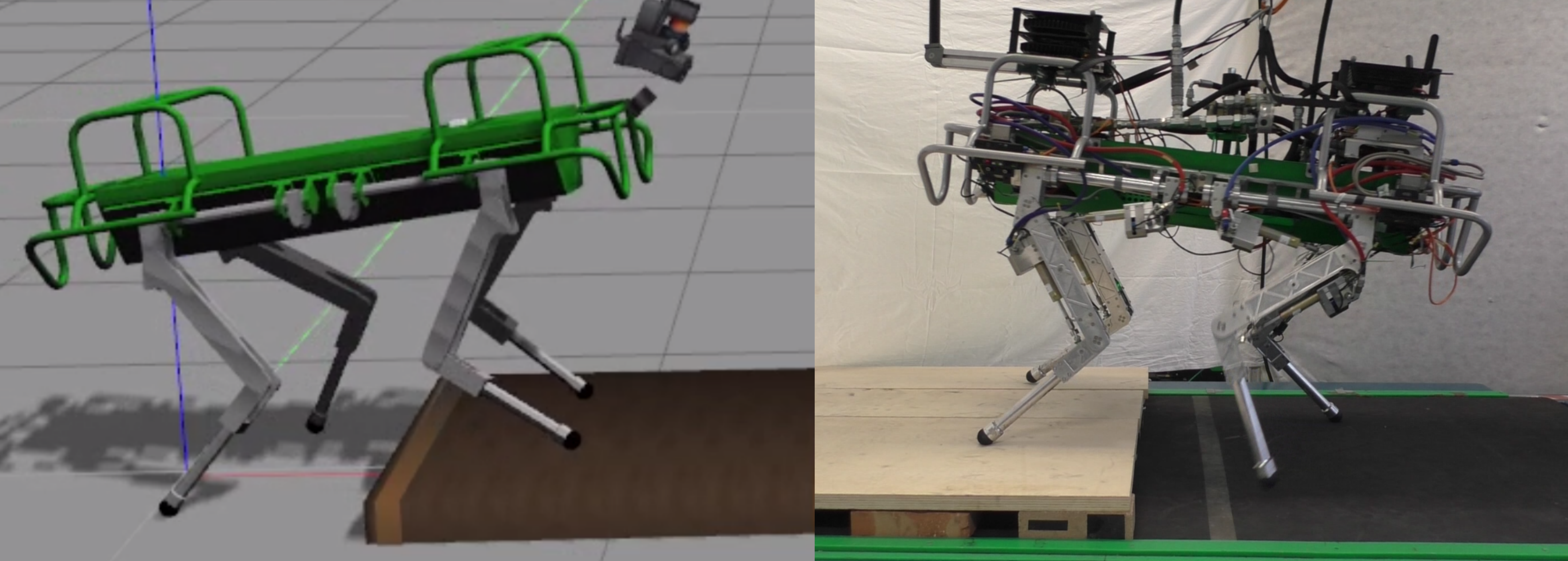}
	\caption{HyQ robot stepping up a pallet of $10cm$ in both simulation (Gazebo) and hardware experiment.}
	\label{fig:titlePic}
\end{figure}
Exploiting the constraint that we described in Section \ref{sec:shinCollision}, \gls{hyq} was able to walk for $1$m, performing three cycles of crawl in $11$s to step onto $15$cm high pallet in simulation and onto $10$cm high pallet on the hardware robot.  Fig. \ref{fig:shinCollisionSim} shows the base position $x$ (continuous line) and tracking error (dashed line) with respect to the desired trajectory computed by the planner in three following different versions:
\subsubsection{Zero Point Foot and No Shin Collision Avoidance (red lines)}
This corresponds to the formulation given in \cite{Winkler2018} in which both shin collision and the foot size are neglected. The algorithm took $\sim 50$s to find an optimal solution. In the upper plot of Fig. \ref{fig:shinCollisionSim} we can see that, in this case, the tracking error (dashed line) grows until the experiment is stopped because the robot falls down. As it can been seen in the attached video\footnote{\href{https://www.youtube.com/watch?v=bpjlRvtVwe8}{\texttt{https://www.youtube.com/watch?v=bpjlRvtVwe8}}} the robot collides with the corner of the edge, due to a non-robust choice of the foothold. The attached video shows that even a terrain with a relatively low obstacle ($10$cm) cannot be overcome without explicitly considering the feet and leg geometry.
\subsubsection{Non-Zero Point Foot Size and No Shin Collision Avoidance (green lines)}
\begin{figure}
		\centering
		\includegraphics[width=\linewidth]{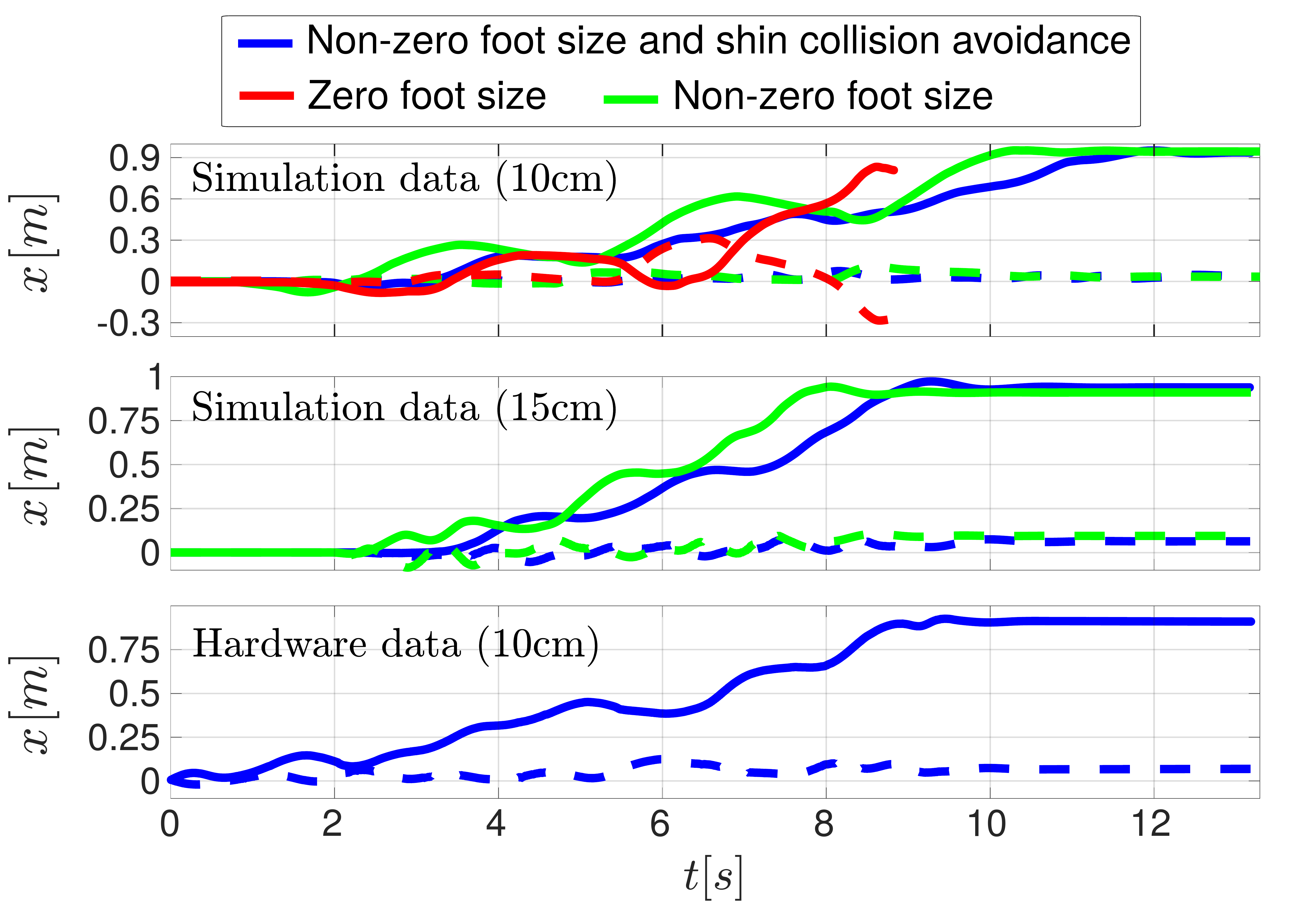}
		\caption{Shin collision planning: tracking performances of simulations and experiments with three different terrain constraints (\textit{zero foot size, non-zero foot size} and \textit{non-zero foot size and shin}) for a walk of $1$m with three crawl cycles. The thick lines represent the base position $x$ while the dashed lines represent the tracking error with respect to the desired trajectory along the same $x$ coordinate.}
		\label{fig:shinCollisionSim}
	\end{figure}

In this case we enforce in the planner a foot radius $r$ of $2$cm, while we do not include any shin collision avoidance constraint. The computation time increased by 30 \% compared to the first scenario ($\sim 70$s) in the case of $10$cm high pallet. For the $15$cm high pallet the solver took $105$s to find an optimal solution. This constraint guarantees the successful navigation in the non flat terrain, but comparing the upper and the middle plot of  Fig. \ref{fig:shinCollisionSim} it can be seen that increasing of the height of the step will also increase the tracking error.
\subsubsection{Non-Zero Point Foot Size and Shin Collision Avoidance (blue lines)}
This version corresponds to the constraint described in Section \ref{sec:shinCollision}. In case of a $10$cm high pallet, the algorithm took twice as long as the first simulation ($\sim 100$s), while a larger computational time ($130$s) was required in case of a $15$cm high pallet, due to the more challenging terrain and due to the absence of a proper initialization. According to the the characteristics of \gls{hyq}, we have chosen $s$ = 0.3m, $\beta$=\ang{37} for the hind legs and $\beta$=\ang{127} for the front legs. The shin collision is thus possible on \gls{hyq} only with hind legs when walking up a step and front legs when walking down a step; which is automatically captured by the definition of the constraint defined in (\ref{eq:shin_collision}). For this experiment we checked possible collision for the knee and for two points. Unlike the previous version of the planner, in this case the tracking error did not increase for a higher step, thanks to the larger robustness given by the shin collision avoidance constraint. 
\section{Conclusion}

In this paper we presented two theoretical contributions consisting of feasibility constraints aimed at increasing the robustness of trajectories that are optimized using the \gls{srbd} model. The first constraint focuses on including the joint-torque limits constraint and approximates the way these limits are mapped into admissible contact forces depending on the leg's configuration. The proposed approximation is suitable for robots having contact points, such as quadrupeds and for motion planning applications based on simplified models, such as \gls{srbd} or \gls{cd}. The second constraint, instead, is able to describe and approximate the volume of the robot's legs in such a way to avoid undesired collisions between the lower limbs and the environment. 

The experimental contribution of this paper consists of the hardware deployment of the optimal trajectories obtained with the formulation of \cite{Winkler2018} which would not have been possible on non-flat terrain without the feasibility constraints that we proposed above.

Future works include the development of strategies for online replanning (the solver optimizes while the robot walks) and the usage of pre-computed feasible solutions for the warm start of every new nonlinear trajectory optimization to reduce the computational time. 

\bibliographystyle{IEEEtran}
\bibliography{library}

\end{document}